\documentclass[10pt,twocolumn,letterpaper]{article}

\usepackage[pagenumbers]{cvpr}

\usepackage{cuted} %

\usepackage{currfile} %

\usepackage{caption} %

\usepackage{multirow}
\usepackage{array}
\usepackage{float}

\definecolor{cvprblue}{rgb}{0.21,0.49,0.74}
\usepackage[breaklinks,colorlinks,allcolors=cvprblue]{hyperref}

\newcommand{\method}{\textsc{VLG-Drive}}

\newcommand{\suppnum}[1]{\ref{#1}}
\newcommand{\suppref}[1]{Appendix~\suppnum{#1}}

\title{VLG-Drive: Unifying Vision-Language and Geometric Priors for \\Autonomous Driving}

\author{%
Junge Zhang\textsuperscript{1,*}\quad
Jie Zhu\textsuperscript{2,*}\quad
Yanqi Jia\textsuperscript{3}\quad
Xiangyu Xu\textsuperscript{4}\quad
Zixun Huang\textsuperscript{4}\\[1pt]
Wenbin He\textsuperscript{4}\quad
Xin Li\textsuperscript{4}\quad
Jiachen Li\textsuperscript{1}\quad
Liu Ren\textsuperscript{4}\\[4pt]
{\normalsize \textsuperscript{1}Georgia Institute of Technology\quad
\textsuperscript{2}University of North Carolina at Chapel Hill}\\
{\normalsize \textsuperscript{3}City University of New York\quad
\textsuperscript{4}Bosch North America}\\[2pt]
{\small \textsuperscript{*}Equal contribution}%
}

\begin{document}
\maketitle
\begin{strip}
    \centering
    \includegraphics[width=\textwidth]{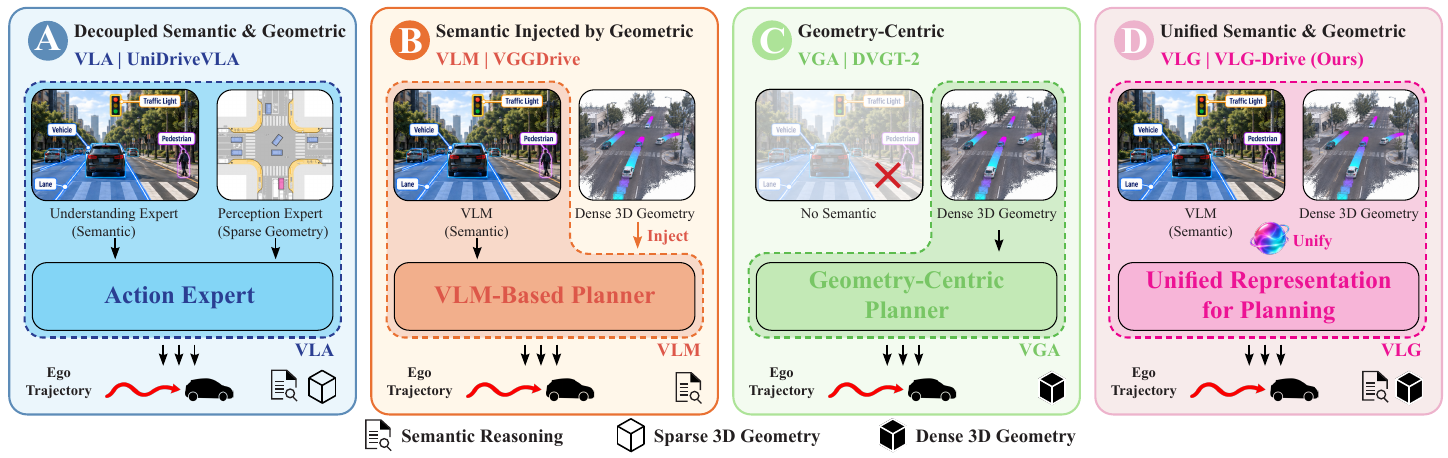}
    \captionsetup{hypcap=false}
    \captionof{figure}{\textbf{Semantic-geometric interfaces for action planning.} (A) UniDriveVLA decouples semantics and geometry into an understanding expert and a sparse-geometry perception expert that feed an action expert~\citep{li2026unidrivevla}. (B) VGGDrive injects dense 3D geometry into a VLM-based planner~\citep{wang2026vggdrive}. (C) DVGT-2 plans with a geometry-centric model that has no semantic stream~\citep{zuo2026dvgt2}. (D) \method{} unifies VLM semantics and dense 3D geometry in a joint representation for planning.}
    \captionsetup{hypcap=true}
    \label{fig:teaser}
    \vspace{-2mm}
\end{strip}

\begin{abstract}
A driving planner needs both scene semantics, which vision-language models (VLMs) provide, and spatial structure, which geometric foundation models recover from visual history. We present \method{}, which keeps both models frozen and trains only a fusion interface and a planner from trajectory labels. At each camera-grid cell, the interface fuses the paired VLM and geometry features into one joint token through a geometry-anchored residual, and a command-conditioned planner reads these tokens to predict the ego trajectory. \method{} achieves $31.78$ EPDMS on NavHard under perturbed initial states, the highest among the compared methods, and $88.27$ PDMS on NavTest from cameras alone. Pointwise fusion outperforms independent VLM and geometry streams on both benchmarks. In closed-loop driving on 420 HUGSIM scenarios, \method{} outperforms all compared planners. These results indicate that a lightweight planner trained on fused features of frozen vision-language and geometric foundation models is an effective alternative to adapting either model for planning.
\end{abstract}

\section{Introduction}

Driving decisions depend on both what a scene means and what motion its physical structure permits. Recognizing a merging vehicle must be related to its position and the available space; interpreting a turn must account for the road layout. Vision-language models (VLMs) connect visual observations with semantic and task context~\citep{radford2021clip,liu2023llava,bai2025qwen3vl}, while geometric foundation models learn spatial representations through reconstruction~\citep{wang2023dust3r,wang2025vggt}. How can a driving planner make joint use of these different sources?
Recent systems answer this question in different ways (Figure~\ref{fig:teaser}). UniDriveVLA decouples semantics and geometry into an understanding expert and a sparse-geometry perception expert~\citep{li2026unidrivevla}, VGGDrive injects dense 3D geometry into a VLM-based planner~\citep{wang2026vggdrive}, and DVGT-2 plans with a geometry-centric model that has no semantic stream~\citep{zuo2026dvgt2}. A simpler alternative would give the planner both feature streams and let its attention combine them. Both features, however, describe the same image locations, so their correspondence is known in advance, yet this design leaves the planner to rediscover it from limited trajectory labels. What matters is therefore not only access to both sources but also the interface that combines them.

Motivated by these observations, we propose \method{}, which binds VLM and geometry features that share a camera-grid address before the planner reads them. The key idea is to let trajectory supervision shape how the two priors interact, so that scene meaning and spatial structure inform ego motion together. A residual interface anchored on the geometry feature combines each dense VLM--geometry pair into one joint token, which replaces the pair in the planner memory. Fusion is local while planning is global, so the interface decides how the two priors describe each location, and a separate, command-conditioned decoder decides which locations matter for the requested maneuver. Both encoders stay frozen, and trajectories are decoded by a dedicated planner rather than a language model.

We evaluate \method{} primarily on NavHard, because its perturbed initial states require the planner to respond to a changed spatial relationship between the ego vehicle and its surroundings~\citep{cao2025pseudosimulation}. On NavHard, \method{} reaches $31.78$ EPDMS, the highest among the compared methods. On NavTest, which evaluates recorded initial states~\citep{dauner2024navsim}, it reaches $88.27$ PDMS from camera input alone. Pointwise fusion outperforms independent streams at every tested label fraction on both benchmarks, with the largest NavTest margin at $10\%$ of the labels ($82.23$ versus $79.21$ PDMS). The same interface improves a second frozen geometry backbone, DVGT-2, on both benchmarks. In closed-loop driving on 420 HUGSIM scenarios with reactive traffic~\citep{zhou2024hugsim}, \method{} scores higher than LTF and well above VGGDrive and DVGT-2. A component-level analysis shows that vision-language context mainly improves drivable-area compliance after the perturbation with both geometry backbones, whereas collision avoidance is already supported by the geometric representation.

The contributions are threefold: \textbf{(i)} \method{}, a geometry-anchored residual interface that fuses frozen VLM and geometry features into joint tokens for a dedicated trajectory planner; \textbf{(ii)} evidence that the interface, not access to both priors alone, drives the gain: pointwise fusion outperforms independent streams at every tested label fraction and improves two geometry backbones; and \textbf{(iii)} the highest NavHard EPDMS among the compared methods, the highest closed-loop HD-Score among the compared planners on HUGSIM, and strong three-camera NavTest results with frozen foundation models, together with an analysis of where vision-language context helps.

\vspace{-1mm}
\section{Related Work}

\noindent{\textbf{Vision-language models for driving and planning.}}
End-to-end planners connect scene representations to action through bird's-eye-view aggregation~\citep{li2022bevformer}, task queries~\citep{hu2023uniad}, vectorized scenes~\citep{jiang2023vad}, sensor fusion~\citep{chitta2022transfuser}, and the design of perception--planning connections~\citep{weng2024paradrive}. Foundation models add pretrained knowledge to this design space. Image-text alignment, visual instruction tuning, and the Qwen-VL family provide representations for language-conditioned scene understanding~\citep{radford2021clip,liu2023llava,bai2025qwen25vl,bai2025qwen3vl}. Flamingo and BLIP-2 learn interfaces between pretrained visual and language components~\citep{alayrac2022flamingo,li2023blip2}, while OpenVLA adapts pretrained visual features to robotic action~\citep{kim2024openvla}. Driving methods built on VLMs connect these capabilities to planning in several ways. LMDrive uses language-conditioned closed-loop control~\citep{shao2023lmdrive}; DriveGPT4 combines explanations and control prediction~\citep{xu2023drivegpt4}; DriveLM and DriveVLM organize reasoning through structured questions or hierarchical plans~\citep{sima2023drivelm,tian2024drivevlm}. EMMA represents scene elements and trajectories in a common language space~\citep{hwang2024emma}, and AutoVLA generates discrete trajectory actions autoregressively~\citep{zhou2025autovla}. ReCogDrive connects VLM driving priors to a continuous diffusion planner~\citep{li2025recogdrive}. Closest to the representation question studied here, UniDriveVLA coordinates understanding, sparse perception, and action through specialized experts~\citep{li2026unidrivevla}. These VLM-based planners read the scene mainly through 2D image features, which leave the metric position of semantic cues implicit. \method{} instead combines dense geometry features with VLM features before a dedicated trajectory decoder reads them, so that each semantic feature is tied to the 3D structure at its location and the planner knows where that evidence lies in space.

\noindent{\textbf{Geometric foundation models.}}
Reconstruction provides supervision for reusable spatial features. DUSt3R predicts point maps without requiring known camera calibration or poses~\citep{wang2023dust3r}, and MASt3R adds descriptors for image matching~\citep{leroy2024mast3r}. Spann3R and CUT3R accumulate scene information sequentially~\citep{wang2024spann3r,wang2025cut3r}; Fast3R processes multiple views jointly~\citep{yang2025fast3r}, and MonST3R addresses dynamic scenes~\citep{zhang2024monst3r}. VGGT jointly estimates cameras, depth, point maps, and tracks~\citep{wang2025vggt}, and VGGT-$\Omega$ extends geometric pretraining across static and dynamic scenes and studies transfer to vision-language-action models~\citep{wang2026vggtomega}. By reconstructing depth, camera poses, and point maps accurately from images alone, these models let an agent locate scene elements at concrete positions in 3D space rather than inferring their location from appearance. DVGT-2 brings this capability to driving by jointly predicting geometry and trajectories in a streaming architecture without an explicit VLM branch~\citep{zuo2026dvgt2}. VGGDrive instead incorporates dense geometric features into VLM states and lets the VLM generate the trajectory~\citep{wang2026vggdrive}. \method{} combines frozen geometric and VLM representations cell by cell on a shared camera grid in a trajectory-supervised interface and plans without reconstructing an explicit scene.

\noindent{\textbf{Geometry priors in multimodal language models.}}
Several recent models inject geometric features into multimodal language models. Spatial-MLLM builds on Qwen2.5-VL~\citep{bai2025qwen25vl} and merges its 2D visual features with those of a spatial encoder initialized from VGGT~\citep{wang2025vggt} through a lightweight MLP connector~\citep{wu2025spatialmllm}. VG LLM, a video--3D geometry large language model, adds VGGT geometry features to the visual tokens of Qwen2.5-VL before the language model~\citep{zheng2025vgllm}. VLM-3R fuses reconstruction tokens from CUT3R~\citep{wang2025cut3r} with 2D appearance features through cross-attention for spatial and temporal reasoning~\citep{fan2026vlm3r}. In all of these models, the fused features are routed back into the language model, which is tuned with instruction-style supervision and produces its output as text. Most of the VLM-based driving planners discussed above follow the same paradigm. \method{} departs from it. It fuses geometric and vision-language features outside the language model and learns a dedicated planner directly on the fused features, which regresses continuous ego motion from trajectory labels. Neither source model is fine-tuned, and producing a plan requires no language decoding.

\vspace{-1mm}
\section{VLG-Drive: Unifying Vision-Language and Geometric Priors}
\label{sec:method}

\subsection{Planning from complementary priors}

Well-pretrained foundation models already capture much of what a driving planner needs. A driving-adapted VLM reads the semantics of a scene and the task it implies, and a geometric foundation model recovers the scene's spatial structure from visual history. Both have been trained at a scale that downstream driving data cannot match. Recognizing this potential, we ask whether a planner can be learned from their features alone while both foundation models stay frozen. \method{} answers this question by unifying the two priors at the representation level.

\begin{figure*}[t]
    \centering
    \includegraphics[width=\textwidth]{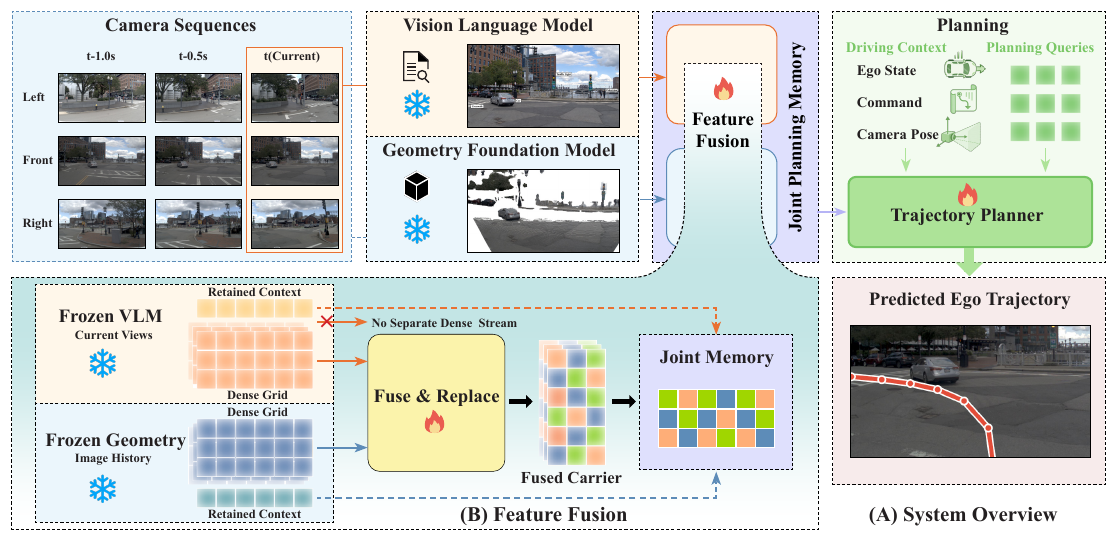}
    \caption{\textbf{The \method{} architecture.} (A) The frozen VLM reads the current views (orange box), and the frozen geometry model reads each camera's image sequence (blue panel). Their features are fused into a joint planning memory, which the trajectory planner reads with planning queries conditioned on the route command, together with ego-state and camera-pose queries, to predict the ego trajectory. Snowflakes mark frozen modules and flames mark trainable ones. (B) At each of the $3\times17\times30$ paired cells, pointwise residual fusion adds a learned update from both sources to the geometry feature (Equation~\ref{eq:fusion}), and the resulting token of the fused carrier replaces both dense features in the joint memory. The retained context of each source (dashed arrows) enters the joint memory directly.}
    \label{fig:overview}
    \vspace{-2mm}
\end{figure*}

Let $\mathcal{I}_t$ denote the current multi-camera observation, $\mathcal{H}_t$ the visual history including the current observation, $u_t$ the route command, and $m_t$ the recent ego-motion history included in the VLM prompt. Frozen encoders produce VLM features $V_t$ and geometry features $G_t$, a trainable interface $F_\theta$ combines them into a joint planning representation $Z_t$, and a dedicated planner $P_\phi$ predicts the future trajectory:
\begin{equation}
\begin{aligned}
    V_t&=E_{\mathrm{vlm}}(\mathcal{I}_t,u_t,m_t), &\qquad G_t&=E_{\mathrm{geo}}(\mathcal{H}_t),\\
    Z_t&=F_\theta(V_t,G_t), &\qquad \hat{\tau}_t&=P_\phi(Z_t,u_t).
\end{aligned}
\label{eq:composition}
\end{equation}
Only $F_\theta$ and $P_\phi$ are trained (Section~\ref{sec:objective}). The route command conditions the VLM prompt and the planner, and motion history enters only through the prompt. This factorization separates representation formation, which decides how the two priors describe each part of the scene, from trajectory decoding, which decides which parts matter for the requested maneuver. Figure~\ref{fig:overview} summarizes the architecture.

\subsection{Forming a joint planning representation}
\label{sec:fuse}

\noindent{\textbf{Establishing correspondence.}}
The two models are pretrained for different objectives, so matching channel dimensions alone does not make their features compatible; the camera image instead provides a common indexing domain. The VLM features are the final-layer hidden states of its language model at the image-token positions, so they are conditioned on the prompt with the route command and motion history. The interface pools both sources to a common $17\times30$ grid per camera and projects them to the planner width $D$. Each grid cell is indexed by $i=(c,h,w)$, where $c\in\{1,2,3\}$ selects the front-left, front, or front-right camera, and $h\in\{1,\dots,17\}$ and $w\in\{1,\dots,30\}$ are the row and column of the cell in that camera's grid. Source-specific adapters produce
\begin{equation}
    v_i=A_v(V_i;\pi_i), \qquad g_i=A_g(G_i;\pi_i),
\end{equation}
where $V_i$ and $G_i$ are the pooled features of cell $i$ from $V_t$ and from the current-frame grid of $G_t$, and $\pi_i$ encodes the viewing ray of the cell. The address specifies \emph{where} the two sources are paired, while the adapters learn \emph{how} each source is represented for planning.

\noindent{\textbf{Pl\"ucker ray encoding.}}
Grid indices alone do not tell the planner where a cell looks, because the same row and column point in different directions in different cameras. Each cell is therefore tagged with its viewing ray in the ego frame. Let $\mathbf{p}_{hw}$ be the homogeneous image coordinate of the cell center, $K_c$ the intrinsic matrix of camera $c$, and $R_c$ and $\mathbf{o}_c$ its rotation and center in the ego frame. The ray is described by its Pl\"ucker coordinates
\begin{equation}
    \mathbf{d}_i=\frac{R_cK_c^{-1}\mathbf{p}_{hw}}{\lVert R_cK_c^{-1}\mathbf{p}_{hw}\rVert}, \qquad
    \mathbf{m}_i=\mathbf{o}_c\times\mathbf{d}_i,
\end{equation}
where the unit direction $\mathbf{d}_i$ gives the orientation of the ray and the moment $\mathbf{m}_i$ distinguishes parallel rays from different camera centers. These coordinates are mapped to the planner width with fixed random Fourier features,
\begin{equation}
    \pi_i=\big[\sin(2\pi B^{\top}\mathbf{x}_i);\,\cos(2\pi B^{\top}\mathbf{x}_i)\big],
    \quad \mathbf{x}_i=\big[\mathbf{d}_i;\,\mathbf{m}_i/s\big],
    \label{eq:ray}
\end{equation}
where $B\in\mathbb{R}^{6\times D/2}$ is a fixed Gaussian matrix and $s=2$\,m scales the moment to the size of the camera rig. Both adapters add this parameter-free encoding (\suppref{sec:implementation}), so the two sources describe each cell in the same ray coordinates.

\noindent{\textbf{Pointwise residual fusion.}}
With both sources aligned at every cell, the remaining question is how to combine them. Passing the two features to the planner as separate tokens would leave it to discover, from trajectory labels alone, which VLM feature belongs to which geometric evidence. Since the grid address already specifies this correspondence, we combine each pair at its address before the planner reads the scene. The combination takes a residual form, with the geometry feature, which carries the metric structure of the cell, as the anchor and a learned update that injects vision-language context:
\begin{equation}
    r_i = f_\theta\!\left([\operatorname{LN}_g(g_i);\operatorname{LN}_v(v_i)]\right), \qquad z_i = g_i+r_i,
    \label{eq:fusion}
\end{equation}
where $f_\theta$ is a nonlinear projection shared across locations and $[\cdot;\cdot]$ denotes channel-wise concatenation. The residual depends on both inputs and thus learns a cross-source interaction. We call the anchoring stream the \emph{carrier}; the VLM carrier swaps the roles of $g_i$ and $v_i$ (Section~\ref{sec:ablation}). Zero-initializing the final projection gives $z_i=g_i$ at the start of training, so optimization begins from the geometric part.

\noindent{\textbf{Fuse and replace.}}
The complete representation is
    $Z_t=\{z_i\}_{i\in\mathcal{S}}\cup\mathcal{C}_t,$
where $\mathcal{S}$ indexes the paired dense grid and $\mathcal{C}_t$ retains the camera and register tokens of the geometry model and two global VLM tokens, the final-layer hidden state averaged over the non-image prompt tokens and the hidden state at the last prompt position, none of which has a dense counterpart. The dense VLM stream thus enters the planner only through the joint tokens, which reduces the planner memory from $3{,}114$ tokens for independent streams to $1{,}584$ in the main setting.

\subsection{Action-conditioned trajectory prediction}
\label{sec:objective}

A dedicated trajectory planner translates $Z_t$ into future ego motion. It follows query-based transformer decoding~\citep{carion2020detr}, but unlike object queries, which form an unordered set, each query $q_k$ corresponds to an ordered future step $k$ and incorporates the route command $u_t$. Self-attention among the queries coordinates the future steps into a consistent trajectory, and cross-attention lets each step retrieve the scene evidence it needs from $Z_t$. Each decoder layer $\ell\in\{1,2,3\}$ predicts incremental ego motion $\hat{\delta}^{(\ell)}_k=(\Delta x,\Delta y,\Delta\psi)$ for eight future steps at $0.5$-second intervals, and the final-layer increments are accumulated to form $\hat{\tau}_t$.

\noindent{\textbf{Training objective.}}
Let $\bar{\delta}_k$ denote the ground-truth increment at step $k$, normalized per channel with training-set statistics, and let $m_k\in\{0,1\}$ mask future steps without ground truth. The trajectory loss of decoder layer $\ell$ is a masked Smooth-$L_1$ loss on the normalized increments of the three channels $j\in\{x,y,\psi\}$,
\begin{equation}
    \mathcal{L}^{(\ell)}_{\mathrm{traj}}
    =\frac{1}{\sum_{k}m_k}\sum_{k=1}^{8}m_k\sum_{j}\operatorname{SmoothL1}\!\big(\hat{\delta}^{(\ell)}_{k,j}-\bar{\delta}_{k,j}\big).
\end{equation}
Additional ego-state, ego-pose, and per-camera queries in the decoder contribute three terms, $\mathcal{L}_{\mathrm{state}}$, $\mathcal{L}_{\mathrm{pose}}$, and $\mathcal{L}_{\mathrm{cam}}$, whose targets come from the same driving log and from the frozen geometry encoder (\suppref{sec:implementation}). The total objective applies all terms at every decoder layer,
\begin{equation}
    \mathcal{L}=\sum_{\ell=1}^{3}\lambda_\ell\Big(
    \mathcal{L}^{(\ell)}_{\mathrm{traj}}
    +0.2\,\mathcal{L}^{(\ell)}_{\mathrm{state}}
    +0.5\,\mathcal{L}^{(\ell)}_{\mathrm{pose}}
    +0.1\,\mathcal{L}^{(\ell)}_{\mathrm{cam}}\Big),
    \label{eq:loss}
\end{equation}
with layer weights $(\lambda_1,\lambda_2,\lambda_3)=(0.2,0.6,1.0)$. Supervision therefore comes entirely from the logged ego motion and the frozen encoders' own outputs.

\section{Experiments}
\label{sec:experiments}

\subsection{Experimental protocol}

We train on NAVSIM v1 NavTrain and evaluate on two benchmarks. NavTest contains $12{,}146$ scenarios scored by PDMS under the official four-second non-reactive protocol~\citep{dauner2024navsim}; PDMS multiplies no-at-fault collision and drivable-area compliance terms with a weighted average of ego progress, time to collision, and comfort. NavHard contains $5{,}912$ observations scored by the official two-stage EPDMS~\citep{cao2025pseudosimulation}. Stage 1 scores the plan from $450$ recorded scenes; Stage 2 scores plans from $5{,}462$ synthetic start states rendered around the human driver's position four seconds later, each weighted by its proximity to the endpoint of the Stage-1 plan, and EPDMS multiplies the two stage scores of each scene. \method{} builds its frozen feature branches on a driving-adapted 2B Qwen3-VL model, the Stage-1 checkpoint released with UniDriveVLA~\citep{li2026unidrivevla}. The VLM receives the current front-left, front, and front-right images, and VGGT-$\Omega$ receives three timestamps per camera (\suppref{sec:history}). Only the interface and planner are trained, $106.1$M parameters in total. The planner is a three-layer decoder of width 1024 with eight attention heads and eight planning queries, trained from scratch for $24{,}000$ updates with batch size 128, AdamW, a learning rate of $2.5\times10^{-4}$, and $4{,}000$ warmup updates on cached encoder outputs. Published baseline scores come from the cited sources, and we evaluate VGGDrive and DVGT-2 from their official checkpoints (\suppref{sec:baselines}); \suppref{sec:implementation} lists all recipes, and \suppref{sec:cost} reports training and inference cost. Each entry is one training run; \suppref{sec:seeds} reports the variation across runs.

\subsection{Overall planning performance}
\label{sec:overall}

\noindent{\textbf{Perturbed-state planning.}}
NavHard tests whether a planner produces useful motion when the ego state departs from the recorded trajectory. Table~\ref{tab:navhard} reports every component at both stages and compares \method{} with published baselines and with the two closest alternatives to our design, VGGDrive, which feeds geometry into a VLM that generates the plan, and DVGT-2, which plans from geometry alone.
\begin{table*}[t]
\caption{\textbf{Planning from perturbed initial states on NavHard.} Scores are percentages; $\uparrow$ marks metrics where higher is better. Bold marks the best value in each column within each stage. S1/S2 are the evaluation stages, S.\ is the score of each stage, and EPDMS is the official combined score. NC, DAC, DDC, TLC, EP, TTC, LK, HC, and EC denote no-at-fault collisions, drivable-area and driving-direction compliance, traffic-light compliance, ego progress, time to collision, lane keeping, history comfort, and extended comfort. Scores for LTF~\citep{dauner2024navsim}, DiffusionDrive~\citep{liao2024diffusiondrive}, GTRS-DP~\citep{li2025gtrs}, and GuideFlow~\citep{liu2025guideflow} are reported by \citet{li2026metis}; we evaluate VGGDrive~\citep{wang2026vggdrive} and DVGT-2~\citep{zuo2026dvgt2}. -- denotes an unavailable result.}
\label{tab:navhard}
\centering
\footnotesize
\setlength{\tabcolsep}{2.5pt}
\begin{tabular*}{\textwidth}{@{\extracolsep{\fill}}llrrrrrrrrrrr}
\toprule
Method & Stage & NC$\uparrow$ & DAC$\uparrow$ & DDC$\uparrow$ & TLC$\uparrow$ & EP$\uparrow$ & TTC$\uparrow$ & LK$\uparrow$ & HC$\uparrow$ & EC$\uparrow$ & S.$\uparrow$ & EPDMS$\uparrow$ \\
\midrule
\multirow{2}{*}{LTF} & S1 & \textbf{97.3} & 80.2 & 97.8 & 99.3 & 83.4 & \textbf{96.2} & 92.9 & \textbf{97.8} & 71.1 & 61.3 & \\
 & S2 & 79.4 & 69.0 & 85.6 & 98.5 & 83.8 & 76.7 & 47.9 & \textbf{97.0} & 70.6 & 39.2 & 24.40 \\
\multirow{2}{*}{DiffusionDrive} & S1 & 96.8 & 86.0 & 98.8 & 99.3 & 84.0 & 95.8 & \textbf{96.7} & 97.6 & \textbf{79.6} & 66.7 & \\
 & S2 & 80.1 & 72.8 & 84.4 & 98.4 & 85.9 & 76.6 & 46.4 & 96.3 & \textbf{72.8} & 40.5 & 27.50 \\
\multirow{2}{*}{GTRS-DP} & S1 & 94.7 & 78.8 & 96.1 & 99.5 & 83.0 & 94.4 & 92.0 & 97.5 & 72.8 & -- & \\
 & S2 & 80.3 & 74.4 & 84.9 & 98.0 & 81.9 & 78.8 & 45.4 & 96.7 & 70.1 & -- & 23.80 \\
\multirow{2}{*}{GuideFlow} & S1 & 96.6 & 80.5 & 96.3 & 99.3 & 82.3 & 94.9 & 91.5 & 97.7 & 67.8 & -- & \\
 & S2 & \textbf{87.3} & \textbf{76.7} & \textbf{88.8} & \textbf{99.2} & 84.3 & \textbf{85.1} & \textbf{49.7} & 93.1 & 44.5 & -- & 27.10 \\
\multirow{2}{*}{VGGDrive} & S1 & 95.4 & 80.9 & 97.6 & 99.1 & \textbf{86.6} & 94.4 & 92.9 & \textbf{97.8} & 68.9 & 68.3 & \\
 & S2 & 76.5 & 70.3 & 82.4 & 97.9 & 89.7 & 72.8 & 49.3 & 95.8 & 58.6 & 36.8 & 26.09 \\
\multirow{2}{*}{DVGT-2} & S1 & 96.0 & \textbf{88.4} & 97.6 & \textbf{100.0} & 85.5 & 96.0 & 95.3 & \textbf{97.8} & 71.1 & \textbf{76.5} & \\
 & S2 & 76.9 & 70.1 & 77.5 & 98.7 & \textbf{92.0} & 72.5 & 47.5 & 82.0 & 38.8 & 35.3 & 28.80 \\
\midrule
\multirow{2}{*}{\textbf{\method{}}} & S1 & \textbf{97.3} & 85.6 & \textbf{99.6} & 99.8 & 84.9 & 95.8 & \textbf{96.7} & 97.6 & 70.7 & 74.9 & \\
 & S2 & 82.4 & 71.7 & 85.7 & 98.5 & 89.4 & 79.3 & 47.4 & 95.9 & 55.7 & \textbf{41.0} & \textbf{31.78} \\
\bottomrule
\end{tabular*}
    \vspace{-1mm}
\end{table*}
\method{} reaches $31.78$ EPDMS, $2.98$ points above the strongest baseline, DVGT-2. The gain comes from Stage 2, which scores the plans made after perturbation. Its stage score of $41.0$ is the highest reported, driven by higher NC and TTC than DVGT-2 at a small cost in ego progress.

\noindent{\textbf{Standard-distribution planning.}}
NavTest evaluates the same planner from recorded initial states, measuring its driving quality under the standard distribution (Table~\ref{tab:navtest}).
\begin{table*}[t]
\caption{\textbf{System-level NAVSIM v1 NavTest planning.} Scores are percentages; $\uparrow$ marks metrics where higher is better, and bold marks the best value in each column. Cameras count physical views, and auxiliary supervision lists the targets other than the trajectory used in downstream training. E2E, VGA, and VLA denote conventional end-to-end, geometry-based, and vision-language-action planners. Published scores are taken from the cited sources, and we evaluate VGGDrive (\suppref{sec:baselines}).}
\label{tab:navtest}
\centering
\footnotesize
\setlength{\tabcolsep}{2pt}
\begin{tabular*}{\textwidth}{@{\extracolsep{\fill}}lllrcrrrrrr}
\toprule
Method & Family & \shortstack{Downstream auxiliary\\supervision} & Cameras & LiDAR & NC$\uparrow$ & DAC$\uparrow$ & TTC$\uparrow$ & Comfort$\uparrow$ & EP$\uparrow$ & PDMS$\uparrow$ \\
\midrule
UniAD~\citep{hu2023uniad} & E2E & Map \& bbox & 8 & -- & 97.80 & 91.90 & 92.90 & \textbf{100.00} & 78.80 & 83.40 \\
Latent TransFuser~\citep{dauner2024navsim} & E2E & Map \& bbox & 3 & -- & 97.40 & 92.80 & 92.40 & \textbf{100.00} & 79.00 & 83.80 \\
TransFuser~\citep{chitta2022transfuser} & E2E & Map \& bbox & 3 & \checkmark & 97.78 & 92.63 & 92.89 & 99.98 & 78.88 & 83.88 \\
PARA-Drive~\citep{weng2024paradrive} & E2E & Map \& bbox & 8 & -- & 97.90 & 92.40 & 93.00 & 99.80 & 79.30 & 84.00 \\
DiffusionDrive~\citep{liao2024diffusiondrive} & E2E & Map \& bbox & 3 & \checkmark & 98.20 & 96.20 & 94.70 & \textbf{100.00} & 82.20 & 88.10 \\
LFG~\citep{strong2026lfg} & VGA & Pseudo 3D \& motion & 1 & -- & 98.20 & 93.70 & 94.40 & \textbf{100.00} & 79.10 & 85.20 \\
DVGT-2~\citep{zuo2026dvgt2} & VGA & Dense geometry & 8 & -- & 97.80 & 97.20 & 93.90 & \textbf{100.00} & 83.40 & 88.60 \\
DVGT-2-NAVSIM~\citep{zuo2026dvgt2} & VGA & Dense geometry & 8 & -- & \textbf{98.70} & \textbf{97.90} & \textbf{95.80} & \textbf{100.00} & \textbf{84.30} & \textbf{90.30} \\
AutoVLA~\citep{zhou2025autovla} & VLA & CoT reasoning & 3 & -- & 96.89 & 92.42 & 88.06 & 99.94 & 75.82 & 80.54 \\
ReCogDrive~\citep{li2025recogdrive} & VLA & Driving QA & 3 & -- & 98.30 & 95.10 & 94.30 & \textbf{100.00} & 81.10 & 86.80 \\
VGGDrive~\citep{wang2026vggdrive} & VLA & None & 3 & -- & 96.14 & 93.44 & 91.17 & 99.93 & 80.29 & 84.02 \\
\textbf{\method{}} & VLA & Ego state \& camera pose & 3 & -- & 98.55 & 96.16 & 95.20 & \textbf{100.00} & 82.27 & 88.27 \\
\bottomrule
\end{tabular*}
\end{table*}
\method{} reaches $88.27$ PDMS from three cameras, above DiffusionDrive, which additionally uses LiDAR, and $4.25$ points above VGGDrive, which generates trajectories with its VLM. DVGT-2 trains all parameters of its geometry backbone and planner on large-scale driving data and reaches $88.60$ PDMS, and DVGT-2-NAVSIM further fine-tunes the full model on NAVSIM with eight cameras to reach $90.30$. \method{} keeps both foundation models frozen, with a geometry backbone not trained for driving, and trains only $106.1$M parameters in about $58$ GPU-hours, against about ten days on 64 GPUs for DVGT-2 (\suppref{sec:cost}); it comes within $0.33$ PDMS of DVGT-2 and exceeds the same checkpoint by $2.98$ EPDMS on NavHard, which makes a lightweight planner on general-purpose priors a promising route for driving.
\begin{table}[t]
\caption{\textbf{Closed-loop driving on HUGSIM} (420 scenarios). Scores in percent; Rc is route completion, and brackets give the 95\% scenario-bootstrap interval of HD-Score. LTF is the NAVSIM-trained client provided with HUGSIM.}
\label{tab:closedloop}
\centering
\setlength{\tabcolsep}{2.5pt}
\resizebox{\linewidth}{!}{%
\begin{tabular}{lrrrrrl}
\toprule
Method & NC$\uparrow$ & DAC$\uparrow$ & TTC$\uparrow$ & Comfort$\uparrow$ & Rc$\uparrow$ & HD-Score$\uparrow$ \\
\midrule
LTF~\citep{dauner2024navsim} & 43.0 & 85.2 & 38.8 & 93.4 & 35.7 & 21.2 [18.2, 24.3] \\
VGGDrive~\citep{wang2026vggdrive} & 49.7 & 71.1 & 39.8 & 97.3 & 23.2 & 10.7 [\phantom{0}9.1, 12.5] \\
DVGT-2~\citep{zuo2026dvgt2} & 30.1 & 82.5 & 27.0 & \textbf{99.4} & 33.6 & 16.1 [13.2, 19.1] \\
\midrule
\method{} & \textbf{59.2} & \textbf{89.2} & \textbf{51.6} & 98.2 & \textbf{40.6} & \textbf{25.9} [22.9, 28.9] \\
\bottomrule
\end{tabular}}
    \vspace{-2mm}
\end{table}
\begin{table*}[t]
\caption{\textbf{Fusion and carrier configurations.} NavTest components and PDMS, and NavHard EPDMS, in percent; $\uparrow$ marks metrics where higher is better, and bold marks the best value in each column. DDC is driving-direction compliance; other metrics follow Table~\ref{tab:navtest}. All rows use the base recipe (width 512, $12{,}000$ updates; Table~\suppnum{tab:recipes}).}
\label{tab:fusion}
\centering
\scriptsize
\setlength{\tabcolsep}{3pt}
\resizebox{0.85\textwidth}{!}{%
\begin{tabular}{llrrrrrrrr}
\toprule
Dense representation & Interaction & NC$\uparrow$ & DAC$\uparrow$ & DDC$\uparrow$ & TTC$\uparrow$ & Comfort$\uparrow$ & EP$\uparrow$ & PDMS$\uparrow$ & EPDMS$\uparrow$ \\
\midrule
Independent streams & Decoder attention only & 98.15 & 95.34 & 98.23 & 94.63 & \textbf{100.00} & 81.08 & 87.12 & 28.69 \\
VLM carrier & No dense geometry injection & 97.39 & 94.47 & 98.11 & 93.18 & 99.98 & 79.70 & 85.23 & 25.89 \\
VLM carrier & Pointwise fusion & 98.42 & 95.88 & \textbf{98.28} & 94.85 & \textbf{100.00} & 81.85 & 87.78 & 29.47 \\
Geometry carrier & No dense VLM injection & 98.19 & 95.23 & 98.08 & 94.33 & 99.99 & 81.22 & 86.87 & 29.18 \\
Geometry carrier & Pointwise fusion & \textbf{98.59} & \textbf{95.92} & 98.24 & \textbf{95.14} & 99.99 & \textbf{82.11} & \textbf{88.06} & \textbf{30.75} \\
\bottomrule
\end{tabular}}
    \vspace{-2mm}
\end{table*}

\subsection{Closed-loop driving on HUGSIM}
\label{sec:closedloop}

NAVSIM scores a plan without letting it change the scene. We therefore also drive the NAVSIM-trained planners in closed loop on HUGSIM~\citep{zhou2024hugsim}, which renders reconstructed scenes from nuScenes~\citep{caesar2019nuscenes}, Waymo~\citep{sun2020waymo}, KITTI-360~\citep{liao2022kitti360}, and PandaSet~\citep{xiao2021pandaset} with reactive inserted vehicles. The ego vehicle follows each plan for $0.25$\,s per step, for up to 400 steps. Table~\ref{tab:closedloop} reports the HUGSIM metrics, and \suppref{sec:closedloop-details} lists the inputs of each planner. HD-Score combines route completion with per-step collision, drivable-area, time-to-collision, and comfort terms, and we give its 95\% scenario-bootstrap interval.
Paired over all 420 scenarios, \method{} exceeds LTF by $4.7$ HD-Score points ($95\%$ interval $[1.1, 8.4]$), VGGDrive by $15.2$ ($[12.3, 18.2]$), and DVGT-2 by $9.8$ ($[6.3, 13.4]$). \method{} has the highest collision, drivable-area, time-to-collision, and route-completion scores of all compared planners; only DVGT-2 scores higher on comfort. The two foundation-model baselines have different weaknesses: VGGDrive, which decodes trajectories with its VLM, has the lowest drivable-area and route-completion scores, and DVGT-2, which plans from geometry alone, has the lowest collision and time-to-collision scores. \method{}, which fuses both priors for a dedicated planner, shows neither weakness, and its largest margins over DVGT-2 are in collision and time-to-collision, the terms that also drive its NavHard advantage.

\subsection{Interaction between pretrained representations}
\label{sec:ablation}

The system comparison does not reveal whether the gain comes from the fusion interface or simply from giving the planner both sources. Table~\ref{tab:fusion} therefore compares independent streams with the VLM and geometry carriers of Section~\ref{sec:fuse}; the no-injection variants keep only the partner's global context, which separates injecting the partner from compacting the memory.

Geometry-carrier fusion reaches $88.06$ PDMS and $30.75$ EPDMS, compared with $87.12$ and $28.69$ for independent streams, with higher NC, DAC, TTC, and EP. The full-data arms of Table~\ref{tab:data-efficiency} and a control without the motion-history prompt (\suppref{sec:nohist}) show the same ordering. Fuse-and-replace keeps one dense token per cell instead of two, so its benefit could come from compaction alone. The no-injection geometry carrier tests this: with the same memory layout but no VLM injection, it scores $86.87$ PDMS and $29.18$ EPDMS, compared with $87.12$ and $28.69$ for independent streams, whereas filling that memory with joint tokens raises PDMS to $88.06$, with the largest gains in EP, TTC, and DAC, and EPDMS to $30.75$. Compaction thus accounts for none of the NavTest gain and for $0.49$ of the $2.06$-point NavHard gain. For the VLM carrier, fusion likewise raises PDMS from $85.23$ to $87.78$ and EPDMS from $25.89$ to $29.47$, and the geometry carrier scores higher in both settings on both benchmarks, which motivates anchoring the residual on geometry. Correct pairing also matters at test time: shifting the VLM grid of the main model by one cell, so that each VLM feature meets the geometry of its horizontal neighbor, lowers PDMS by $2.53$ and EPDMS by $4.14$ (\suppref{sec:calib}). Access to both sources is therefore not enough; they should interact before readout, at the same grid location and with geometry as the anchor.

\begin{table}[t]
\caption{\textbf{Planning with reduced downstream labels.} Columns give the share of NavTrain samples used for downstream training. The subsets are nested and consist of whole driving logs, with $7{,}215$, $18{,}209$, and $72{,}902$ samples from $135$, $301$, and $1{,}152$ logs. NavTest PDMS and NavHard EPDMS are percentages; $\uparrow$ marks metrics where higher is better, and bold marks the best value in each column. All arms use the main training recipe.}
\label{tab:data-efficiency}
\centering
\setlength{\tabcolsep}{4pt}
\resizebox{\linewidth}{!}{%
\begin{tabular}{lrrrrrr}
\toprule
 & \multicolumn{2}{c}{10\%} & \multicolumn{2}{c}{25\%} & \multicolumn{2}{c}{100\%} \\
\cmidrule(lr){2-3}\cmidrule(lr){4-5}\cmidrule(lr){6-7}
Representation & PDMS$\uparrow$ & EPDMS$\uparrow$ & PDMS$\uparrow$ & EPDMS$\uparrow$ & PDMS$\uparrow$ & EPDMS$\uparrow$ \\
\midrule
Geometry only (G) & 80.21 & \textbf{21.54} & 84.33 & 24.00 & 86.97 & 27.99 \\
Independent streams (I) & 79.21 & 18.96 & 84.62 & 23.53 & 87.27 & 28.43 \\
Pointwise fusion (F) & \textbf{82.23} & 21.51 & \textbf{85.43} & \textbf{24.03} & \textbf{88.27} & \textbf{31.78} \\
\bottomrule
\end{tabular}}
\par\bigskip
\caption{\textbf{Alternative geometry backbones.} NavTest PDMS and NavHard EPDMS in percent; bold marks the better value per backbone. G is geometry only and F is pointwise fusion; the \mbox{VGGT-$\Omega$} F row is the main model.}
\label{tab:backbone-transfer}
\centering
\footnotesize
\setlength{\tabcolsep}{8pt}
\begin{tabular}{llrr}
\toprule
Backbone & Rep. & PDMS$\uparrow$ & EPDMS$\uparrow$ \\
\midrule
\multirow{2}{*}{VGGT-$\Omega$} & G & 86.97 & 27.99 \\
 & F & \textbf{88.27} & \textbf{31.78} \\
\midrule
\multirow{2}{*}{DVGT-2} & G & 86.66 & 22.87 \\
 & F & \textbf{87.48} & \textbf{23.78} \\
\bottomrule
\end{tabular}
    \vspace{-2mm}
\end{table}

\begin{table}[t]
\caption{\textbf{Fusing features from different encoder layers.} NavTest PDMS and NavHard EPDMS in percent; bold marks the best value per column within each block. Geometry taps are zero-indexed, and VLM blocks are one-indexed decoder layers (\suppref{sec:depth-protocol}).}
\label{tab:depth}
\centering
\small
\setlength{\tabcolsep}{4pt}
\resizebox{\linewidth}{!}{%
\begin{tabular}{lllrr}
\toprule
Setting & Geo.\ taps & VLM & PDMS$\uparrow$ & EPDMS$\uparrow$ \\
\midrule
\multicolumn{5}{l}{\textit{Geometry depth, final-layer VLM}} \\
Final-layer fusion & O23 & Final & 88.27 & \textbf{31.78} \\
CAT3 & O4/O17/O23 & Final $\times3$ & \textbf{88.31} & 31.34 \\
DUP3 & O23 $\times3$ & Final $\times3$ & 88.14 & 31.64 \\
\midrule
\multicolumn{5}{l}{\textit{Cross-source layer pairing}} \\
LP-LATE & O17/O23 & 28/28 & 88.09 & 28.06 \\
LP-PAIR & O17/O23 & 14/28 & \textbf{88.44} & \textbf{30.56} \\
\bottomrule
\end{tabular}}
\end{table}

\begin{figure*}[t]
    \centering
    \includegraphics[width=\textwidth]{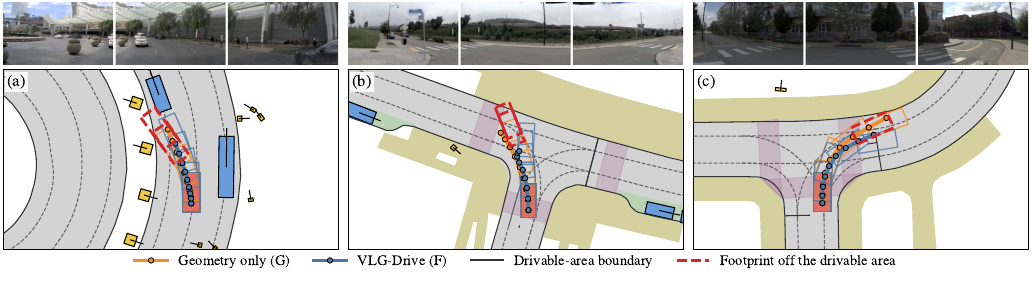}
    \vspace{-6mm}
    \caption{\textbf{Perturbed NavHard starts where vision-language context keeps the plan on the road.} Each column shows a Stage-2 start: the three front views (top) and a bird's-eye view (bottom) with the plans of the VGGT-$\Omega$ geometry-only planner (G, orange) and \method{} (F, blue). Outlines show the simulated ego footprint, and red dashed ones leave the drivable area: G cuts the inside of a curve (a) and the corner of an intersection (b), and swings wide when leaving a turn (c), whereas F stays on the road.}
    \label{fig:navhard-examples}
    \vspace{-1mm}
\end{figure*}

\subsection{Scope of the representation effect}
\label{sec:scope}

Table~\ref{tab:fusion} establishes the benefit of fusion with the full training set and one geometry backbone. We next test whether it persists across different amounts of downstream trajectory supervision and different geometry backbones, the two axes that determine how widely the approach applies.

\noindent{\textbf{Across downstream supervision levels.}}
\label{sec:data-efficiency}
If the joint representation makes better use of the pretrained priors, its advantage should be most visible when downstream labels are scarce. Table~\ref{tab:data-efficiency} tests this by comparing geometry alone (G), both sources as independent streams (I), and their pointwise fusion (F), each trained on $10\%$, $25\%$, or $100\%$ of the NavTrain samples (\suppref{sec:transfer-protocol}).
Fusion scores above independent streams at every label fraction on both benchmarks. Its NavTest margin is largest at $10\%$ of the labels ($82.23$ versus $79.21$ PDMS), where independent streams score below geometry-only memory on both benchmarks: with $7{,}215$ samples, the planner cannot yet exploit separate VLM tokens, whereas fused tokens raise PDMS by $2.02$ over geometry alone. On NavHard, fusion and geometry-only memory remain close at $10\%$ and $25\%$, while the full-data main model reaches $31.78$ EPDMS, $3.79$ points above geometry only. On NavTest, fusion thus extracts the most from the frozen priors when labels are scarce; on NavHard, its advantage over geometry alone grows with supervision.

\noindent{\textbf{Across geometry backbones.}}
\label{sec:backbone-transfer}
A useful composition should hold beyond a single geometry checkpoint. Table~\ref{tab:backbone-transfer} therefore tests whether adding VLM features through the same interface also helps DVGT-2, a frozen geometry backbone that differs from VGGT-$\Omega$ in both architecture and training data. Because DVGT-2 encodes undistorted images, the VLM grid is resampled into its camera frame so that both sources share grid addresses (\suppref{sec:transfer-protocol}).
Fusion improves both backbones on both benchmarks, by $1.30$ PDMS and $3.79$ EPDMS for VGGT-$\Omega$ over its full-data geometry-only arm in Table~\ref{tab:data-efficiency}, and by $0.82$ and $0.91$ in the DVGT-2 pair. The fusion interface thus carries over to a second geometry backbone, so its benefit is not specific to the architecture or training data of VGGT-$\Omega$.

\subsection{Fusing features from different encoder layers}
\label{sec:depth}

Both foundation models expose a hierarchy of internal features, while the main system fuses only the final layer of each. Since intermediate layers retain finer spatial detail, Table~\ref{tab:depth} tests whether fusing several levels helps. CAT3 fuses early, intermediate, and final geometry taps, and DUP3 repeats the final tap through three adapters to separate depth diversity from the number of feature banks; a second comparison pairs the geometry taps O17 and O23 with VLM decoder blocks, where LP-LATE uses the final block 28 for both taps and LP-PAIR pairs O17 with the intermediate block 14.
Final-layer fusion, the main model, achieves the best NavHard EPDMS among the geometry-depth variants and matches the best NavTest PDMS to within $0.04$ points, and CAT3 and DUP3 perform similarly. Pairing an intermediate VLM block with the intermediate geometry tap raises PDMS by $0.35$ and EPDMS by $2.50$ over using the final VLM block for both taps. Final-layer fusion is therefore a simple and competitive design, as no multi-level variant improves both metrics over it, and multi-level fusion benefits from pairing VLM and geometry layers by depth.

\subsection{Where vision-language context helps}
\label{sec:analysis}

\begin{table}[t]
\caption{\textbf{NavHard components with and without vision-language context.} G is the VGGT-$\Omega$ geometry-only planner of Table~\ref{tab:backbone-transfer} and F the fused main system. NC, DAC, DDC, TTC, and EP follow Table~\ref{tab:navhard}, and S.\ is the stage score. Scores are percentages; bold marks the better value in each stage. Table~\suppnum{tab:vlm-contribution} in \suppref{sec:vlm-components} lists all components for both geometry backbones.}
\label{tab:vlm-components}
\centering
\footnotesize
\setlength{\tabcolsep}{4pt}
\begin{tabular}{llrrrrrr}
\toprule
Stage & Rep. & NC$\uparrow$ & DAC$\uparrow$ & DDC$\uparrow$ & TTC$\uparrow$ & EP$\uparrow$ & S.$\uparrow$ \\
\midrule
\multirow{2}{*}{S1} & G & \textbf{97.78} & 84.44 & 99.33 & \textbf{96.22} & 84.15 & 74.06 \\
 & F & 97.33 & \textbf{85.56} & \textbf{99.56} & 95.78 & \textbf{84.92} & \textbf{74.87} \\
\midrule
\multirow{2}{*}{S2} & G & 82.42 & 68.56 & 85.25 & 77.88 & 87.76 & 38.74 \\
 & F & \textbf{82.43} & \textbf{71.74} & \textbf{85.70} & \textbf{79.31} & \textbf{89.39} & \textbf{40.99} \\
\bottomrule
\end{tabular}
\end{table}

The preceding studies show that VLM features improve planning, but not which driving abilities each prior provides. We therefore compare the VGGT-$\Omega$ geometry-only planner of Table~\ref{tab:backbone-transfer} with the fused main system on NavHard (Table~\ref{tab:vlm-components}). The gain concentrates in Stage 2, after the ego state is perturbed, where the stage score rises by $2.25$ points compared with $0.81$ in Stage 1. The largest Stage-2 improvement is in drivable-area compliance ($+3.18$), while no-at-fault collisions are unchanged ($+0.01$). Vision-language context thus keeps the displaced planner on the road (Figure~\ref{fig:navhard-examples}), whereas geometry already supports collision avoidance; the two priors contribute different driving abilities, which is why combining them outperforms geometry alone. With DVGT-2, fusion likewise concentrates its gain in Stage 2 ($+1.53$ versus $+0.62$) and improves drivable-area compliance (\suppref{sec:vlm-components}). \suppref{sec:interventions} further shows that the planner reads a small set of fused cells along the driving path.

\raggedbottom
\section{Conclusion and Limitations}

\method{} turns frozen vision-language and geometry features into a trajectory-supervised joint planning representation. It reaches $31.78$ EPDMS on NavHard, the highest among the compared methods, and $88.27$ PDMS on NavTest from camera input alone. Pointwise fusion outperforms independent streams at every tested label fraction and improves two different geometry backbones on both benchmarks. In closed-loop driving on HUGSIM, \method{} outperforms all compared planners. Vision-language context mainly improves drivable-area compliance after perturbation with both geometry backbones, while the geometric representation already supports collision avoidance. A lightweight planner trained on the fused features of frozen vision-language and geometric foundation models is therefore an effective alternative to adapting either model for planning.
Because \method{} keeps both foundation models frozen and trains only a lightweight interface and planner, fully fine-tuning the geometry model on larger driving datasets, as DVGT-2 does, could further improve the results given sufficient compute. Closed-loop evaluation on further benchmarks such as nuPlan~\citep{karnchanachari2024nuplan} and further pretrained encoder combinations will test how broadly joint representation learning benefits driving planners.

{
    \small
    \bibliographystyle{ieeenat_fullname}
    \bibliography{references}
}

\clearpage
\appendix
\setcounter{table}{0}
\setcounter{figure}{0}
\renewcommand{\thetable}{S\arabic{table}}
\renewcommand{\thefigure}{S\arabic{figure}}
\def\theHtable{supp.\arabic{table}}
\def\theHfigure{supp.\arabic{figure}}
\section{Implementation Details}
\label{sec:implementation}

\noindent{\textbf{Sources of vision-language and geometric priors.}}
The vision-language branch uses a driving-adapted Qwen3-VL model with 2B parameters, the publicly released Stage-1 nuScenes checkpoint of UniDriveVLA~\citep{li2026unidrivevla}. According to its released training configuration, it is fine-tuned from Qwen3-VL-2B-Instruct~\citep{bai2025qwen3vl} on driving question-answering and planning corpora built on nuScenes~\citep{caesar2019nuscenes} and on BDD-X, DRAMA, LingoQA, CODA-LM, Talk2Car, SUTD-TrafficQA, and MAPLM, together with a general visual question-answering subset. None of these sources is derived from nuPlan~\citep{karnchanachari2024nuplan}, from which the NAVSIM logs are drawn, so no NAVSIM training or test log reaches the frozen VLM during its adaptation. The geometry branch uses VGGT-$\Omega$. Both encoders are frozen during downstream training. VLM features correspond to the current front-left, front, and front-right images. The main geometry features are extracted per camera with visual history, and the current-frame grid is passed to the planner.

\noindent{\textbf{Dense layout and spatial indexing.}}
Both branches supply a $17\times30$ grid per camera. Qwen3-VL encodes each $1920\times1080$ view at its native resolution, which its processor rounds to $1920\times1088$; with $16$-pixel patches and $2\times2$ token merging, each view yields a $34\times60$ grid of merged visual tokens. We take the final hidden states of the language model at these image-token positions, restore their two-dimensional layout from the processor's grid size, and average-pool each $2\times2$ block, so every cell of the $17\times30$ grid summarizes a fixed $64\times64$-pixel image region. VGGT-$\Omega$ features of the same view are pooled to the same grid. In an earlier front-camera, geometry-only study, a coarser $8\times11$ grid scored $86.74$ NavTest PDMS against $87.37$ for $17\times30$. Pooling and camera ordering define corresponding row-major addresses, so correspondence is defined at the grid level. Source-specific adapters project the features to the decoder width. Ray directions are computed at cell centers from the recorded intrinsics and camera-to-ego extrinsics, and the Pl\"ucker encoding of Equation~\ref{eq:ray} is applied to both sources. The geometry adapter adds the encoding to its projected feature. The VLM adapter uses a second encoding of the same form with its own random matrix, maps it through a zero-initialized linear layer, and adds it to $V_i$ before projection, so the VLM features are unchanged at initialization. The geometry adapter additionally adds a learned camera embedding and a fixed two-dimensional sinusoidal encoding of the grid position.

\noindent{\textbf{Fusion and memory.}}
The pointwise fusion module is a two-layer MLP with GELU activation, source-specific LayerNorm, and a zero-initialized output layer. Concatenating independent streams retains $1{,}530$ dense VLM tokens and $1{,}530$ dense geometry tokens, together with context, for $3{,}114$ memory tokens. Pointwise fusion retains $1{,}530$ joint tokens, three camera tokens, 48 register tokens, a null token, and two global VLM tokens (the final-layer hidden state averaged over the non-image prompt tokens and the hidden state at the last prompt position), totaling $1{,}584$ decoder memory tokens.

\noindent{\textbf{Training configurations.}}
The main system and the data-efficiency, backbone, feature-level, and history-length studies use the main recipe; Sec.~\ref{sec:fusion-controls} lists the recipe of each fusion-study row. Table~\ref{tab:recipes} compares the two recipes. Both have three decoder layers, eight attention heads, and a feed-forward expansion of four. The main recipe has 106.1M trainable parameters outside the frozen encoders. When the source representation is fixed, frozen-encoder outputs are cached once and reused during downstream training; experiments that vary internal feature levels extract the relevant features online.

\begin{table}[htbp]
\caption{\textbf{Downstream training recipes.} Both keep the pretrained encoders frozen and train the downstream modules from scratch.}
\label{tab:recipes}
\centering
\small
\begin{tabular}{lrr}
\toprule
Setting & Base recipe & Main recipe \\
\midrule
Decoder width & 512 & 1024 \\
Decoder layers & 3 & 3 \\
Training iterations & 12,000 & 24,000 \\
Batch size & 256 & 128 \\
Learning rate & $10^{-3}$ & $2.5\times10^{-4}$ \\
Warmup iterations & 1,000 & 4,000 \\
\bottomrule
\end{tabular}
\end{table}

\noindent{\textbf{Planning queries and losses.}}
Eight planning queries represent future poses at $0.5$-second intervals over four seconds. A learned four-way command embedding (left, straight, right, or unknown) conditions these queries. Each decoder layer applies query self-attention, memory cross-attention, and a feed-forward block. The model predicts normalized per-step $(\Delta x,\Delta y,\Delta\psi)$ in the NAVSIM ego frame and is trained with the objective of Equation~\ref{eq:loss}. Outputs are denormalized per channel and cumulatively summed. The ego-state, ego-pose, and per-camera queries share the decoder layers with the planning queries, and their prediction heads serve only the training objective. $\mathcal{L}_{\mathrm{state}}$ is an $L_1$ loss on the normalized current velocity and acceleration. $\mathcal{L}_{\mathrm{pose}}$ is a Smooth-$L_1$ loss on the ego motion since the previous frame, with the heading change represented by its sine and cosine. $\mathcal{L}_{\mathrm{cam}}$ is a Smooth-$L_1$ consistency loss between the predicted camera-pose encodings and those of the frozen geometry encoder.

\noindent{\textbf{Prompt and causal inputs.}}
The VLM prompt contains the native NAVSIM command and a recent history-trajectory string in the convention expected by the driving model. Motion history reaches the model only through this prompt and the geometry history. The planner receives no ego-state input; ego velocity and acceleration serve only as auxiliary targets. Sec.~\ref{sec:nohist} removes the history-trajectory string.

\section{Additional Representation Studies}
\label{sec:additional}

\subsection{Geometry history length}
\label{sec:history}

\begin{table}[htbp]
\caption{\textbf{Temporal context in the three-camera fusion planner.} The vision-language branch receives the current views; only the number of timestamps supplied per camera to the geometry model changes, and all rows use the main recipe. The three-timestamp row is the main system of Tables~\ref{tab:navhard} and~\ref{tab:navtest}. Scores use the percentage scale; $\uparrow$ marks metrics where higher is better, and bold marks the best value in each column.}
\label{tab:history-length}
\centering
\small
\setlength{\tabcolsep}{6pt}
\begin{tabular}{lrr}
\toprule
Geometry & NavTest & NavHard \\
timestamps & PDMS$\uparrow$ & EPDMS$\uparrow$ \\
\midrule
1 & 83.30 & 28.68 \\
2 & 87.85 & 30.68 \\
3 & \textbf{88.27} & \textbf{31.78} \\
4 & \textbf{88.27} & 30.52 \\
\bottomrule
\end{tabular}
\end{table}

Table~\ref{tab:history-length} shows that visual history is essential for the geometry branch. A second timestamp raises NavTest PDMS from $83.30$ to $87.85$ and NavHard EPDMS from $28.68$ to $30.68$. A third timestamp raises both scores further, to $88.27$ PDMS and $31.78$ EPDMS, the best NavHard result. A fourth timestamp leaves NavTest at $88.27$ and lowers NavHard to $30.52$. Three timestamps, the setting of the main system, give the best overall result.

The first additional timestamp, which supplies the parallax and recent motion that a single frame lacks, accounts for most of the benefit ($4.55$ PDMS and $2.00$ EPDMS), and the third adds $0.42$ PDMS and $1.10$ EPDMS.

\subsection{Fusion without the motion-history prompt}
\label{sec:nohist}

Motion history enters \method{} in two ways: as the history-trajectory string in the VLM prompt and as the earlier timestamps seen by the geometry model. To test whether the fusion gain depends on the explicit motion history, we remove the history-trajectory string from the prompt, both when the VLM features of the training set are cached and at test time, and retrain the fusion (F) and independent-stream (I) arms with the main recipe. The geometry model still receives three timestamps, so the planner retains visual motion cues; the single-timestamp row of Table~\ref{tab:history-length} shows the effect of removing those.

\begin{table}[htbp]
\caption{\textbf{Planning without the motion-history prompt.} NavTest PDMS and NavHard EPDMS in percent for the main-recipe arms trained without the history-trajectory string in the prompt.}
\label{tab:nohist}
\centering
\small
\setlength{\tabcolsep}{4pt}
\begin{tabular}{lrr}
\toprule
Arm & NavTest PDMS$\uparrow$ & NavHard EPDMS$\uparrow$ \\
\midrule
Independent streams (I) & 87.66 & 29.49 \\
Pointwise fusion (F) & 87.85 & 31.84 \\
\bottomrule
\end{tabular}
\end{table}

Without the prompt, fusion still outperforms independent streams on both benchmarks, by $0.19$ PDMS on NavTest and $2.35$ EPDMS on NavHard.

\subsection{Variation across training runs}
\label{sec:seeds}

Each entry in the main-text tables is a single training run. Table~\ref{tab:seeds} reports three training runs of each arm of the fusion study with the main recipe: pointwise fusion (F, the full-data \method{} model), independent streams (I), and geometry only (G). The first run of each arm is the one reported in the main text; the further runs of F use seeds $1$ and $0$, and those of I and G seeds $1$ and $2$.

\begin{table}[htbp]
\caption{\textbf{Results over training runs of the fusion study.} Each arm is trained three times with the main recipe; the first run is the one reported in the main text. NavTest PDMS and NavHard EPDMS in percent.}
\label{tab:seeds}
\centering
\setlength{\tabcolsep}{3pt}
\resizebox{\linewidth}{!}{%
\begin{tabular}{lrrrrrr}
\toprule
 & \multicolumn{2}{c}{Fusion (F)} & \multicolumn{2}{c}{Indep.\ streams (I)} & \multicolumn{2}{c}{Geometry only (G)} \\
\cmidrule(lr){2-3}\cmidrule(lr){4-5}\cmidrule(lr){6-7}
Run & PDMS$\uparrow$ & EPDMS$\uparrow$ & PDMS$\uparrow$ & EPDMS$\uparrow$ & PDMS$\uparrow$ & EPDMS$\uparrow$ \\
\midrule
1 (reported) & 88.27 & 31.78 & 87.27 & 28.43 & 86.97 & 27.99 \\
2 & 88.35 & 31.81 & 87.54 & 30.45 & 86.76 & 27.10 \\
3 & 88.21 & 31.39 & 87.56 & 30.01 & 86.90 & 27.62 \\
\midrule
Mean $\pm$ s.d. & 88.28 $\pm$ 0.07 & 31.66 $\pm$ 0.23 & 87.46 $\pm$ 0.16 & 29.63 $\pm$ 1.06 & 86.88 $\pm$ 0.11 & 27.57 $\pm$ 0.45 \\
\bottomrule
\end{tabular}}
\end{table}

On both benchmarks, every fusion run scores above every independent-streams run, and every independent-streams run above every geometry-only run. Over the three runs, \method{} reaches $31.66 \pm 0.23$ EPDMS on NavHard and $88.28 \pm 0.07$ PDMS on NavTest.

\subsection{NavHard components with and without vision-language context}
\label{sec:vlm-components}

Table~\ref{tab:vlm-contribution} lists every NavHard component for the VGGT-$\Omega$ geometry-only planner of Table~\ref{tab:backbone-transfer} and the fused main system, the two planners compared in Section~\ref{sec:analysis}, and the same comparison for the DVGT-2 pair of Table~\ref{tab:backbone-transfer}. The table values are the official aggregate component scores of each stage.

\begin{table*}[t]
\caption{\textbf{NavHard components with and without vision-language context.} Geometry only (G) is the VGGT-$\Omega$ geometry-only planner of Table~\ref{tab:backbone-transfer}; pointwise fusion (F) is the main system of Tables~\ref{tab:navhard} and~\ref{tab:navtest}. The lower block compares the DVGT-2 G and F planners of the same table. Scores are percentages; $\uparrow$ marks metrics where higher is better, and bold marks the better value within each stage. Metric abbreviations follow Table~\ref{tab:navhard}.}
\label{tab:vlm-contribution}
\centering
\scriptsize
\setlength{\tabcolsep}{3pt}
\begin{tabular}{llrrrrrrrrrr}
\toprule
Stage & Representation & NC$\uparrow$ & DAC$\uparrow$ & DDC$\uparrow$ & TLC$\uparrow$ & EP$\uparrow$ & TTC$\uparrow$ & LK$\uparrow$ & HC$\uparrow$ & EC$\uparrow$ & S.$\uparrow$ \\
\midrule
\multirow{2}{*}{S1} & Geometry only (G) & \textbf{97.78} & 84.44 & 99.33 & \textbf{100.00} & 84.15 & \textbf{96.22} & 96.22 & \textbf{97.78} & 68.00 & 74.06 \\
 & Pointwise fusion (F) & 97.33 & \textbf{85.56} & \textbf{99.56} & 99.78 & \textbf{84.92} & 95.78 & \textbf{96.67} & 97.56 & \textbf{70.67} & \textbf{74.87} \\
\midrule
\multirow{2}{*}{S2} & Geometry only (G) & 82.42 & 68.56 & 85.25 & 98.40 & 87.76 & 77.88 & 44.50 & \textbf{96.65} & \textbf{56.04} & 38.74 \\
 & Pointwise fusion (F) & \textbf{82.43} & \textbf{71.74} & \textbf{85.70} & \textbf{98.47} & \textbf{89.39} & \textbf{79.31} & \textbf{47.40} & 95.87 & 55.68 & \textbf{40.99} \\
\midrule
\multicolumn{12}{l}{\emph{DVGT-2 backbone (Table~\ref{tab:backbone-transfer})}} \\
\multirow{2}{*}{S1} & Geometry only (G) & \textbf{96.56} & 85.11 & 98.56 & \textbf{99.78} & 85.31 & \textbf{95.33} & \textbf{96.22} & \textbf{97.78} & \textbf{70.22} & 72.76 \\
 & Pointwise fusion (F) & 95.78 & \textbf{86.89} & \textbf{98.67} & \textbf{99.78} & \textbf{85.48} & 94.89 & 94.44 & \textbf{97.78} & 64.89 & \textbf{73.38} \\
\midrule
\multirow{2}{*}{S2} & Geometry only (G) & \textbf{73.78} & 62.73 & 76.34 & 97.55 & 94.12 & 68.81 & \textbf{49.61} & 75.72 & \textbf{29.70} & 30.10 \\
 & Pointwise fusion (F) & 73.75 & \textbf{63.62} & \textbf{79.97} & \textbf{97.62} & \textbf{94.25} & \textbf{70.22} & 47.10 & \textbf{75.76} & 26.27 & \textbf{31.63} \\
\bottomrule
\end{tabular}
\end{table*}

\subsection{Fusion-study controls}
\label{sec:fusion-controls}

All five rows of Table~\ref{tab:fusion} use the base recipe; Table~\ref{tab:data-efficiency} repeats the comparison of independent streams and fusion with the main recipe. The no-injection geometry carrier removes the paired dense VLM grid but keeps global VLM context. The no-injection VLM carrier removes dense geometry but keeps geometry camera and register context. Pointwise fusion combines the two features that share a grid address.

\subsection{Data-efficiency and backbone-transfer protocols}
\label{sec:transfer-protocol}

\noindent{\textbf{Reduced downstream labels.}}
The $10\%$ and $25\%$ subsets are nested whole-log subsets of NavTrain with fixed membership, so every scenario of a smaller subset also belongs to the larger ones. Trajectory normalization statistics are computed from the training data of each fraction. All arms use the main recipe with $24{,}000$ updates at batch size 128, so smaller subsets are revisited proportionally more often. The geometry-only arm (G) disables the VLM pathway, the independent-stream arm (I) passes both dense grids to the decoder memory, and the fusion arm (F) uses pointwise fusion. The full-data fusion entry and the VGGT-$\Omega$ fusion entry in Table~\ref{tab:backbone-transfer} both reuse the main model reported in Tables~\ref{tab:navhard} and~\ref{tab:navtest}.

\noindent{\textbf{Second geometry backbone.}}
The DVGT-2 arms use the released frozen DVGT-2 encoder features computed from undistorted images. The VLM dense grid, extracted from the raw images, is transformed into the undistorted camera frame by calibration-derived bilinear feature resampling, and the same corrected intrinsics define the ray descriptors of both sources. The two global VLM tokens are kept unchanged. The geometry-only and fused DVGT-2 planners each train a new downstream head on the full $72{,}902$-example training set with the main recipe.

\subsection{Feature-level controls}
\label{sec:depth-protocol}

\noindent{\textbf{Setup.}}
Table~\ref{tab:depth} uses three cameras, three-frame geometry histories, and a current-frame $17\times30$ grid per camera. Each configuration trains its adapters, fusion interface, and planner from scratch with the main recipe (Table~\ref{tab:recipes}), and CAT3/DUP3 and LP-LATE/LP-PAIR use the same training sample order within each pair.

\noindent{\textbf{Geometry-depth control.}}
The final-layer fusion row reuses the main model of Tables~\ref{tab:navhard} and~\ref{tab:navtest}, which fuses final-layer geometry and VLM features into one bank. The geometry-depth comparison is CAT3 versus DUP3. O4, O17, and O23 denote zero-indexed geometry taps. CAT3 uses an independent geometry adapter for each tap; DUP3 applies three independent adapters to the same raw O23 feature. A shared VLM projection and shared fusion MLP serve all banks. All fused banks enter the shared key/value memory of every decoder layer. Context is retained once. For three banks, dense attention logits receive a $-\log 3$ bias to control the increased dense-versus-context multiplicity. CAT3 and DUP3 have the same number of trainable parameters and the same memory size.

\noindent{\textbf{Cross-source layer-pairing control.}}
LP-LATE and LP-PAIR extract both frozen representations for each sample. The VLM receives $960\times544$ current-view images. Its selected states are image-token outputs of language-decoder blocks 14 and 28. Block 14 is passed through the frozen final RMSNorm; block 28 is already normalized. Geometry uses O17 and O23, where O17 is a deep intermediate tap. LP-LATE uses block 28 for both banks, whereas LP-PAIR uses block 14 with O17 and block 28 with O23. As in CAT3, each bank has its own geometry adapter, and the VLM projection and fusion MLP are shared. Both banks enter every decoder layer's main memory, with a $-\log 2$ dense-logit bias. Late-layer geometry context and global VLM tokens are retained once in both arms.

\section{Evaluation and Diagnostic Protocols}

\subsection{External baselines and aggregation}
\label{sec:baselines}

Published NavTest rows follow the comparison in \citet{wang2026vggdrive}; AutoVLA and ReCogDrive use its supervised-fine-tuning setting. Camera counts in Table~\ref{tab:navtest} denote distinct physical views consumed by each model before any image stitching. The NAVSIM reproductions of UniAD and PARA-Drive use all eight views, whereas Latent TransFuser and TransFuser use the three front-facing views~\citep{dauner2024navsim}; TransFuser additionally uses LiDAR. DiffusionDrive retains the TransFuser camera and LiDAR setup~\citep{liao2024diffusiondrive}. LFG uses a single monocular front camera~\citep{strong2026lfg}. AutoVLA, ReCogDrive, and VGGDrive use the front-left, front, and front-right views~\citep{zhou2025autovla,li2025recogdrive,wang2026vggdrive}. The DVGT-2 NavTest row reports the official result of \citet{zuo2026dvgt2} for its released checkpoint with eight views, the same checkpoint that we evaluate on NavHard. The DVGT-2-NAVSIM row reports their official result after fine-tuning DVGT-2 on NAVSIM with eight views and four frames.

All our baseline evaluations use the official checkpoints released by the authors, without additional training or fine-tuning. For VGGDrive, we run the official NAVSIM checkpoint and inference code on both NavTest and NavHard, with a causal prompt that contains only information available at inference time. For DVGT-2, we evaluate the official base checkpoint on NavHard with eight views, four frames, and official image undistortion.

For external baselines predicting only $(x,y)$, headings are constructed with a forward tangent at non-terminal points and a consistent terminal convention. The NavHard result is read from the official \texttt{extended\_pdm\_score\_combined} summary, which combines the two stages at the scenario level before averaging. Every model in Tables~\ref{tab:fusion}, \ref{tab:data-efficiency}, \ref{tab:depth}, and~\ref{tab:history-length} is scored on all $12{,}146$ NavTest scenarios and $5{,}912$ NavHard observations, and NavTest component values and PDMS are scenario-weighted means.

\subsection{Closed-loop protocol and further results}
\label{sec:closedloop-details}

\noindent{\textbf{Scenarios and simulator.}}
The HUGSIM benchmark defines 436 scenarios. We exclude the same 16 for every method because the released simulator code cannot run them: the 15 PandaSet medium scenarios with index 02, which its constant-velocity planner rejects, and one Waymo hard scenario whose scenario file specifies an invalid start velocity. We made one fix to the released code, an out-of-range read of the route-command list on the final, route-complete step; it affected two episodes, which were rerun. Every method is scored on all 420 remaining scenarios.

\noindent{\textbf{Inputs.}}
The simulator renders the front-left, front, and front-right cameras with the median NAVSIM front-rig calibration, with the front camera placed at the reconstruction's front camera. Planners receive camera history at 2\,Hz; before enough history exists at the start of an episode, it is filled by constant-speed back-extrapolation. LTF runs through HUGSIM's NAVSIM client. DVGT-2 runs its released checkpoint on the three front cameras at three timestamps, the camera setup of the DVGT-2 arms in Table~\ref{tab:backbone-transfer}, because HUGSIM cannot faithfully render the rear and side views of its native eight-camera input for all datasets. VGGDrive runs its official NAVSIM checkpoint on the three current views with the causal prompt of Sec.~\ref{sec:baselines}; an unparseable output falls back to a constant-velocity plan for that step, which happened in 18 of $8{,}561$ steps. The simulator waits for each plan, so planning latency does not affect the outcome.

\noindent{\textbf{Results by dataset.}}
Table~\ref{tab:closedloop-datasets} breaks HD-Score down by source dataset. KITTI-360 is the hardest for every method.

\begin{table}[htbp]
\caption{\textbf{Closed-loop HD-Score by source dataset} (percent; number of scenarios in parentheses).}
\label{tab:closedloop-datasets}
\centering
\small
\setlength{\tabcolsep}{4pt}
\resizebox{\linewidth}{!}{%
\begin{tabular}{lrrrr}
\toprule
Method & nuScenes & Waymo & KITTI-360 & PandaSet \\
 & (88) & (107) & (113) & (112) \\
\midrule
LTF & \textbf{34.5} & 21.7 & 2.3 & 29.3 \\
VGGDrive & 12.0 & 11.9 & 1.7 & 17.7 \\
DVGT-2 & 15.3 & 19.0 & \textbf{7.1} & 23.0 \\
\midrule
\method{} & 29.8 & \textbf{35.0} & 6.7 & \textbf{33.4} \\
\bottomrule
\end{tabular}}
\end{table}

\noindent{\textbf{Run-to-run variation.}}
Closed-loop episodes are not bit-reproducible, because non-deterministic GPU kernels change the first plan by a few centimeters and the loop amplifies the difference. We reran \method{} on 54 scenarios on the same GPU model and on 55 scenarios on a different GPU model. The HD-Score of a single episode varies between runs with a standard deviation of $12$ points on the same GPU model and $16$ points across GPU models, but the mean over the rerun scenarios moves by only $-0.4$ points ($95\%$ interval $[-4.9, 4.1]$) and $+1.4$ points ($[-4.4, 7.3]$), respectively. Every episode of \method{} in Table~\ref{tab:closedloop} ran on the same GPU model (RTX 6000 Ada), so over 420 scenarios this run-to-run noise changes its mean by under one point.

\subsection{Computational cost}
\label{sec:cost}

Table~\ref{tab:cost} summarizes training and inference cost. \method{} trains only its interface and planner on cached encoder outputs; the caches are computed once per source and reused by all runs that share it. Inference runs both frozen encoders online for every step, and the 2B VLM dominates its latency: it accounts for $65$ of the $87$ TFLOPs per step and the geometry encoder for $22$, so the geometry-only planner needs a quarter of the compute and $40\%$ of the latency. \method{} needs slightly more compute per step than VGGDrive but runs $2.7$ times faster, because VGGDrive generates about $99$ tokens per step one after another; each token adds only $0.015$ TFLOPs but needs its own pass through the 7B model. Of the planners we measured, DVGT-2 needs the least compute, and its latency is close to that of the geometry-only planner. Training \method{}, including the one-time caches, takes about $58$ GPU-hours, whereas DVGT-2 reports about ten days on 64 H20 GPUs.

\begin{table}[htbp]
\caption{\textbf{Training and inference cost.} Latency is the median per-step planning time and memory the peak GPU memory in HUGSIM on one RTX 6000 Ada. FLOPs are counted for one planning step on HUGSIM input (three $1920{\times}1080$ cameras and the history each planner uses), over matrix multiplications, convolutions, and attention, with one multiply-add counted as two FLOPs. Total parameters are those loaded for inference. $^\dagger$\,Reported in the respective paper; n/r: not reported.}
\label{tab:cost}
\centering
\small
\setlength{\tabcolsep}{3pt}
\resizebox{\linewidth}{!}{%
\begin{tabular}{lrrrrrr}
\toprule
Method & Total & Trainable & Training & Latency & Memory & TFLOPs \\
 & params & params & (GPU-h) & (ms) & (GiB) & per step \\
\midrule
\method{} & 3.38B & 106.1M & 29$^{a}$ & 1123 & 9.8 & 87.0 \\
Geometry only (G) & 1.25B & 103.0M & 27$^{a}$ & 450 & 5.3 & 21.7 \\
VGGDrive & 9.39B & 8.48B$^{b}$ & n/r$^{b}$ & 3028 & 21.6 & 77.0 \\
DVGT-2 & 1.71B & 1.71B$^{c}$ & $\approx$15{,}000$^{\dagger c}$ & 425 & 11.4 & 15.5 \\
\bottomrule
\end{tabular}}
\par\vspace{2pt}
\parbox{\linewidth}{\footnotesize
$^{a}$\,24{,}000 updates on 8 RTX PRO 6000 GPUs (about 3.6\,h for \method{}; G is about $5\%$ faster per update), plus about 29\,GPU-h, once, to cache the VLM and geometry features.
$^{b}$\,The second stage fine-tunes the VLM and the geometry adapter with VGGT frozen; 7.81B if the vision encoder is also frozen, which the paper does not state. Two plus two epochs on 16 H200 GPUs.
$^{c}$\,The paper does not state which modules are frozen; the released configuration trains all parameters. About ten days on 64 H20 GPUs.}
\end{table}

\subsection{Robustness to calibration errors}
\label{sec:calib}

Both feature grids index the same camera image, so the pairing of VLM and geometry features does not depend on calibration. Calibration enters only through the ray encoding of Equation~\ref{eq:ray}. Table~\ref{tab:calib} perturbs the intrinsics and camera-to-ego extrinsics used by this encoding at test time, without retraining, and, as a direct test of the correspondence, shifts the VLM grid by one cell relative to the geometry grid.

\begin{table}[htbp]
\caption{\textbf{Perturbed calibration and correspondence at test time.} NavTest PDMS and NavHard EPDMS in percent for the main model, without retraining. Each perturbation is drawn per scenario and camera and applied to all input frames.}
\label{tab:calib}
\centering
\small
\setlength{\tabcolsep}{6pt}
\begin{tabular}{lrr}
\toprule
Perturbation & PDMS$\uparrow$ & EPDMS$\uparrow$ \\
\midrule
None & 88.27 & 31.78 \\
Rotation, $1^\circ$ & 87.49 & 30.99 \\
Rotation, $3^\circ$ & 83.01 & 26.84 \\
Translation, 10\,cm & 87.25 & 30.42 \\
Focal length, $\pm5\%$ & 87.87 & 31.88 \\
VLM grid shifted by one cell & 85.74 & 27.64 \\
\bottomrule
\end{tabular}
\end{table}

For every scenario and camera, with a fixed seed, a rotation turns the camera-to-ego extrinsic by the given angle about a random axis, a translation moves the camera by 10\,cm in a random direction, and a focal-length error scales $f_x$ and $f_y$ by $1.05$ or $0.95$ with the principal point fixed. The same perturbation applies to all input frames of the scenario. The grid shift moves each camera's VLM grid by one column, repeating the edge column, so that each VLM cell is paired with its horizontal neighbor in the geometry grid while the calibration stays exact.

Small calibration errors cost about one point or less. A $1^\circ$ rotation lowers PDMS by $0.78$ and EPDMS by $0.79$, a 10\,cm translation by $1.02$ and $1.36$, and a $5\%$ focal-length error changes them by $-0.40$ and $+0.10$. Larger rotations matter: at $3^\circ$ the model loses $5.26$ PDMS and $4.94$ EPDMS, and its NavHard score falls below those of DVGT-2, DiffusionDrive, and GuideFlow in Table~\ref{tab:navhard}. Shifting the VLM grid by a single cell lowers PDMS by $2.53$ and EPDMS by $4.14$, so the planner relies on the cell-wise correspondence between the two sources and not only on their presence.

\subsection{Attention and spatial interventions}
\label{sec:interventions}

\noindent{\textbf{Where the planner reads.}}
Planning queries place $99.94\%$ of their cross-attention on the fused dense grid, which holds $96.6\%$ of the memory tokens. Within the grid, attention concentrates on the region the vehicle will traverse, as a projected reference-path tube occupies $2.19\%$ of the dense cells but receives $12.89\%$ of the attention mass. Splitting the dense attention by camera and averaging within each command group shows that the reading follows the route command. Left-command scenes place $36.9\%$ of the attention on the front-left camera and $23.6\%$ on the front-right camera; right-command scenes reverse this ordering, with $15.1\%$ and $32.8\%$. Straight-command scenes concentrate $59.5\%$ of the attention on the front camera, and scenes without a known command spread it nearly evenly across the three views. The same fused memory is therefore read differently according to the requested maneuver.

\noindent{\textbf{Attended locations drive the prediction.}}
To test whether this attention reflects reliance~\citep{jain2019attention,wiegreffe2019attention}, we replace selected features with the mean dense feature of the same camera in every decoder layer and recompute the trajectories. Replacing the top $5\%$ of attended cells shifts waypoints by $0.717$ m, compared with $0.066$ m for equal-size random cells and $0.0009$ m for the bottom $5\%$. Replacing the path tube itself shifts waypoints by $0.414$ m, compared with $0.043$ m for an equal-size random control. The planner's predictions therefore depend on a small set of localized memory cells concentrated along the driving path.

\noindent{\textbf{Protocol.}}
The analysis uses the \method{} model reported in Tables~\ref{tab:navhard} and~\ref{tab:navtest}. The diagnostic set is drawn from the 3,373 validation samples by taking, within each command group (left, straight, right, and unknown), one scene from each of the low, middle, and high trajectory-error strata, so that it spans every route command and a range of difficulty.

Cross-attention is recovered for all decoder layers, attention heads, and planning queries by replaying the observed attention calls with their masks; the recomputed outputs match the original calls to within $1.91\times10^{-5}$. Quantitative measurements use the native attention grid.

In every decoder layer, selected dense key/value inputs are replaced by the mean dense feature from the same camera. The top and bottom $5\%$ of attended cells are compared with three random selections of the same size. A second intervention targets a one-cell-radius neighborhood of valid, visible projected reference waypoints, using equal-size random controls. Projection uses $(x,y,0)$ and calibrated camera transforms, since the trajectory's third channel is heading. Invisible reference nodes are omitted, and the reference trajectory serves only to define the diagnostic region.

The reported shift is the mean distance between the perturbed and original predicted waypoints. Averaged over scenes, the path tube receives $6.74\times$ its share of cells in attention mass.

\subsection{Qualitative examples and failure cases}
\label{sec:qualitative}

\begin{figure*}[t]
    \centering
    \includegraphics[width=\textwidth]{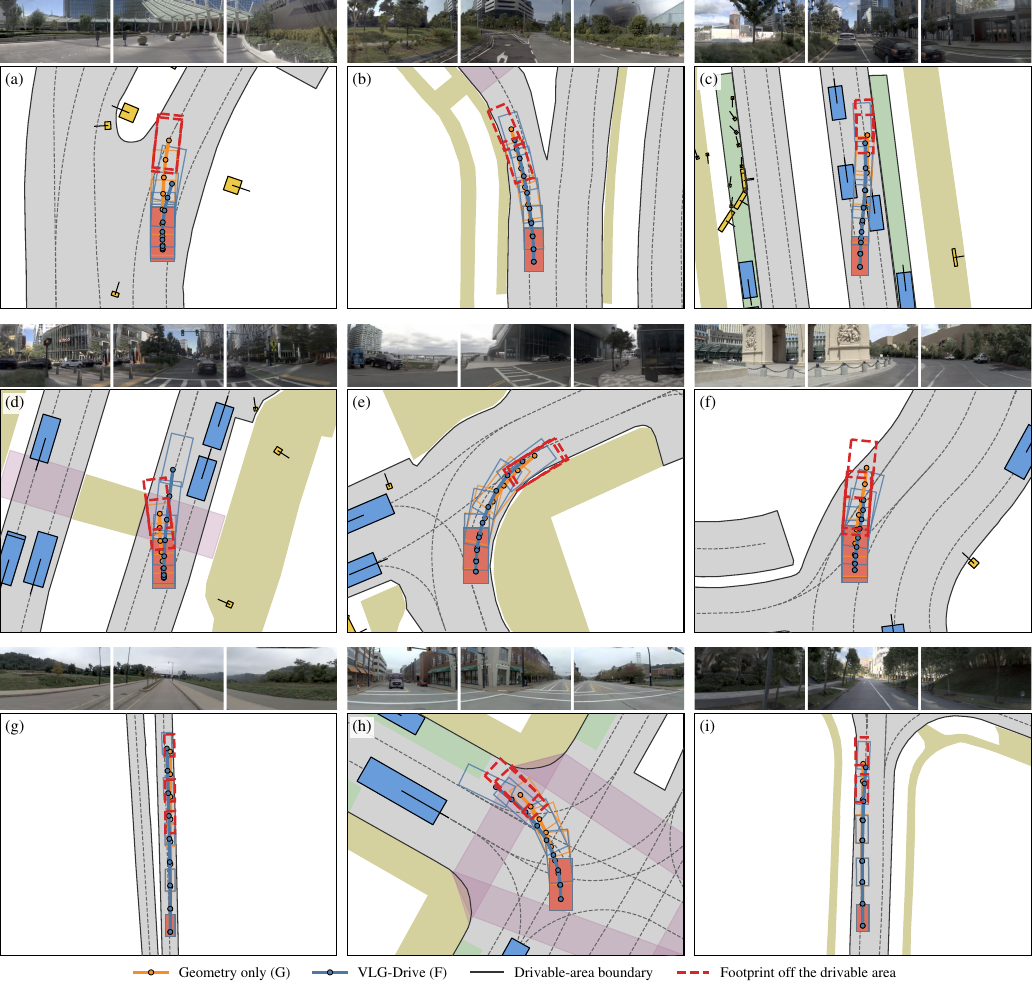}
    \caption{\textbf{Further starts at which only the geometry-only planner leaves the drivable area.} Stage-2 starts of NavHard; the nine sampled starts that are not shown in Figure~\ref{fig:navhard-examples} of the paper. Each panel shows the three front views (top) and a bird's-eye view (bottom) with the plans of G (orange) and \method{} (F, blue). Outlines show the simulated ego footprint, and red dashed ones leave the drivable area (black line). G fails drivable-area compliance at every start and F passes it.}
    \label{fig:supp-examples}
\end{figure*}

\begin{figure*}[t]
    \centering
    \includegraphics[width=\textwidth]{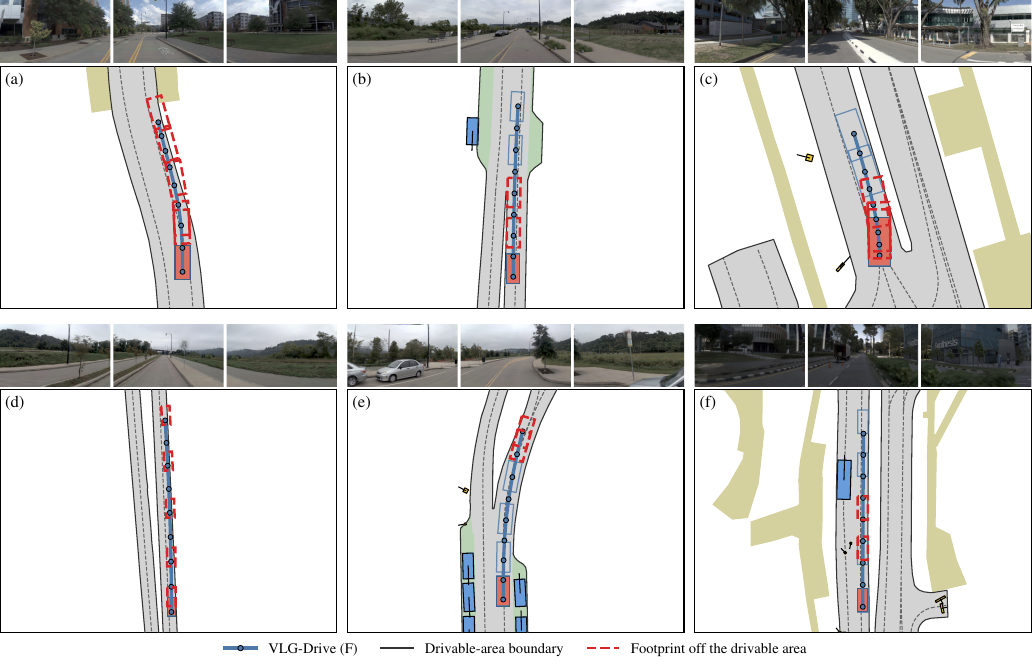}
    \caption{\textbf{Failure cases of \method{} (F).} Six sampled Stage-2 starts of NavHard at which F leaves the drivable area. Each panel shows the three front views (top) and a bird's-eye view (bottom) with the plan of F (blue). Outlines show the simulated ego footprint, and red dashed ones leave the drivable area (black line).}
    \label{fig:supp-failures}
\end{figure*}

\noindent{\textbf{Selection.}}
Drivable-area compliance (DAC) is the lowest of the four multiplicative components of \method{} in Stage 2 of NavHard (NC, DAC, DDC, and TLC in Table~\ref{tab:navhard}), so we look at the plans that leave the drivable area. Figure~\ref{fig:supp-examples} compares the main system (F) with the VGGT-$\Omega$ geometry-only planner (G) of Table~\ref{tab:backbone-transfer} on Stage-2 starts at which DAC is the only one of these components that G fails, F passes all of them, and both planners score above zero from the recorded start of the same scene. A random sample with a fixed seed gives twelve such starts from different scenes; Figure~\ref{fig:navhard-examples} of the paper shows three of them, chosen for legibility, and Figure~\ref{fig:supp-examples} shows the other nine. Figure~\ref{fig:supp-failures} shows failures of F alone, sampled without reference to any other planner: a random sample with a fixed seed of six Stage-2 starts from different scenes at which DAC is the only one of these components that F fails, and at which F scores above zero from the recorded start of the same scene.

\noindent{\textbf{Examples.}}
In Figure~\ref{fig:supp-examples}, G leaves the road at forks and bends (a, b, e, f), at the corner of an intersection (h), and over the edge of a lane (d, g, i), while the plan of F stays inside the drivable area. In (c) a corner of the footprint of G falls into a narrow gap between the lane and the parking area in the map. In several of these starts the two plans nearly coincide, and a small lateral offset at the road edge decides the outcome.

\noindent{\textbf{Failure cases.}}
In Figure~\ref{fig:supp-failures}, the plans continue along the road, and the footprint leaves it sideways. In four starts the footprint begins within $0.2$~m of the edge of the drivable area (b, c, d, f). It crosses that edge in (b), (c), and (f), and the plan then brings it back inside; in (d) it stays over the edge of a narrow lane. In the other two starts the road bends. In (a) the plan runs wide of a left bend and does not return within the horizon, and in (e) the front left corner crosses the outside edge of a right bend at the end of the plan.

\noindent{\textbf{What the failing starts share.}}
Over all Stage-2 starts, those at which F fails DAC differ from those at which it passes in the same two respects. The failing starts begin closer to the edge of the drivable area: the median distance from the edge to the closest corner of the footprint is $1.1$~m, against $1.6$~m. Their routes also bend more: over the next $25$~m, where the route extends that far, the median heading change of its centerline is $19^\circ$, against $4^\circ$. The speed at the start is similar, with medians of $4.1$ and $3.8$~m/s. The distributions overlap, so these are tendencies. The training objective contains no term on the road boundary: its targets are the logged ego motion and the outputs of the frozen encoders (Equation~\ref{eq:loss}). Adding a drivable-area term during training is a natural next step for these failures.

\end{document}